%% file: main.tex
\documentclass[runningheads]{llncs}

\usepackage{eccv}

\usepackage{eccvabbrv}
\usepackage[ruled,vlined]{algorithm2e}
\usepackage{float}
\usepackage{multirow}
\usepackage[table,xcdraw]{xcolor}

\usepackage{graphicx}
\usepackage{booktabs}
\usepackage{tcolorbox}
\usepackage{wrapfig}
\usepackage[accsupp]{axessibility}

\usepackage{hyperref}

\usepackage{orcidlink}

\begin{document}
\let\oldaddcontentsline\addcontentsline
\renewcommand{\addcontentsline}[3]{}
\addtocontents{toc}{\protect\setcounter{tocdepth}{-1}}
\title{Entropy-Gradient Grounding: Training-Free Evidence Retrieval in Vision-Language Models} 

\titlerunning{Entropy-Gradient Grounding}

\author{Marcel Gr\"opl$^{*,} $ \inst{1} \and
 Jaewoo Jung$^{*,} $\inst{3} \and
Seungryong Kim \inst{3}\and
Marc Pollefeys \inst{1}\and
Sunghwan Hong \inst{1,2}}

\authorrunning{M.~Gr\"opl et al.}

\institute{ETH Zurich \and
ETH AI Center\and
KAIST AI\\}

\institute{$^{1}$ETH Z\"urich \quad $^{2}$ETH AI Center \quad $^{3}$KAIST AI \\
\url{https://entropy-gradient-grounding.github.io}}

\maketitle
\let\addcontentsline\oldaddcontentsline
\vspace{-10pt}
\begin{figure}
    \centering
    \includegraphics[width=\linewidth]{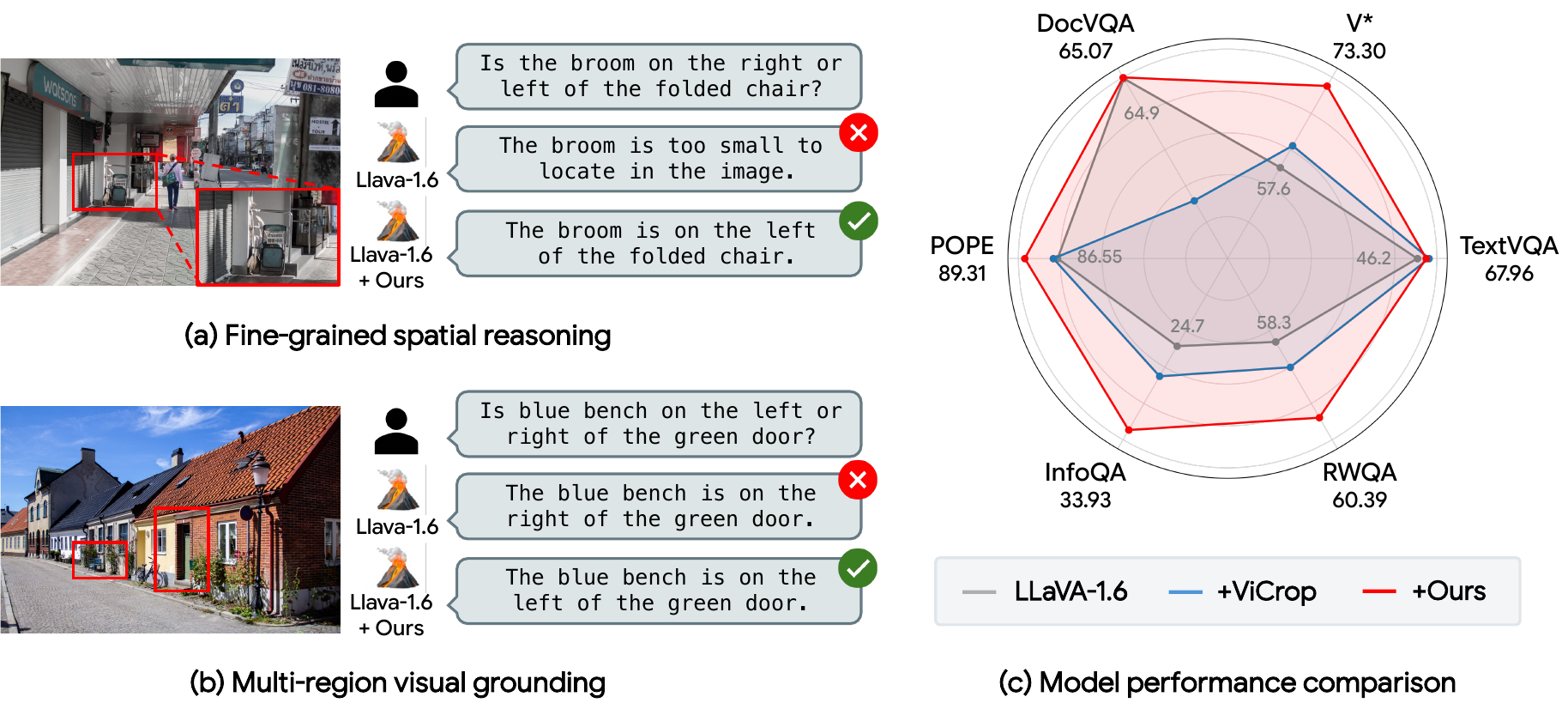}
    \caption{\textbf{Teaser.} As shown in (a) and (b), existing VLMs struggle to answer questions when visual evidence is fine-grained or exists in spatially disjoint regions. We propose a training-free method where we apply a query-based visual grounding method to discover relevant regions and provide these regions as additional image crops, improving performance in both challenging scenarios.}
    \label{fig:teaser}\vspace{-20pt}
\end{figure}

\begin{abstract}
  Despite rapid progress, pretrained vision--language models still struggle when answers depend on tiny visual details or on combining clues spread across multiple regions, as in documents and compositional queries. We address this by framing grounding as test-time evidence retrieval: given a query, the model should actively identify where to look next to resolve ambiguity. To this end, we propose a training-free, model-intrinsic grounding method that uses uncertainty as supervision. Specifically, we compute the entropy of the model’s next-token distribution and backpropagate it to the visual token embeddings to obtain an \emph{entropy-gradient} relevance map, without auxiliary detectors or attention-map heuristics. We then extract and rank multiple coherent regions to support multi-evidence queries, and introduce an iterative zoom-and-reground procedure with a spatial-entropy stopping rule to avoid over-refinement. Experiments on seven benchmarks across four VLM architectures demonstrate consistent improvements over existing methods, with the largest gains on detail-critical and high-resolution settings, while also producing more interpretable evidence localizations.
\end{abstract}

\input{sec/1_Introduction}
\input{sec/2_RelatedWorks}
\input{sec/3_Methods}
\input{sec/4_Experiments}
\input{sec/5_Conclusion}

\bibliographystyle{splncs04}
\bibliography{main}
\clearpage
\appendix
\input{sec/supple}
\end{document}

%% file: sec/1_Introduction.tex
\section{Introduction}
Vision--Language Models (VLMs) have shown impressive performance on a broad range of multimodal tasks, including visual question answering and document understanding \cite{liu2024llavanext,Qwen-VL,wang2025internvl3_5,molmo2024,yoon2025visual,gurbuz2026moving}. Yet many failures persist when the decisive evidence is fine-grained, \textit{e.g.,} small text, symbols, or scattered across disjoint regions, \textit{e.g.,} documents, compositional queries~\cite{Jiang_2025_ICCV,zhang2025mllmsknowlooktrainingfree}, as exemplified in Fig.~\ref{fig:teaser}. These failures suggest that the key challenge is not generation, but \emph{selective perception}: identifying the right visual evidence. Accordingly, we cast such vision--language tasks as \emph{active perception} at inference time: conditioned on a query, the model should perform grounding or decide \emph{where to look} to acquire informative evidence.

A growing body of work explores \emph{training-free} ways to extract grounding signals from pretrained VLMs~\cite{zhang2025mllmsknowlooktrainingfree,kang2025largevisionlanguagemodelneeds}, avoiding additional training that can be prohibitively costly for today’s large-scale models.
Typically, many methods rely on attention maps for numerous downstream tasks~\cite{an2025c3g,han2025emergent,hong2024unifying2,an2025cross}, but converting attention into reliable localization typically requires heuristic choices, \textit{e.g.,} selecting heads/layers and designing post-processing, and the resulting maps can be brittle across backbones and dominated by a single salient region---a poor fit for queries that require aggregating spatially disjoint evidence. A complementary line of work \emph{acts} on these imperfect signals by cropping or zooming into the predicted evidence at test time. ViCrop~\cite{zhang2025mllmsknowlooktrainingfree} exemplifies this direction, showing that zooming can improve downstream accuracy, but its region selection remains heuristic and it typically commits to a \emph{single} crop. As a result, current approaches provide limited support for principled multi-region retrieval and controlled refinement: if the initial saliency is coarse or biased toward one cue, a one-shot zoom may miss secondary evidence, whereas exhaustive grid-based zooming can recover it only at substantial computational cost. 

In this work, we propose a training-free, model-intrinsic grounding approach that treats inference as \emph{test-time evidence acquisition} guided by the model’s own uncertainty. Our key insight is that the entropy of the next-token distribution provides a decision-relevant indicator of visual evidence. Given an image and a query, we compute the entropy of the next-token distribution at a single decoding step and backpropagate it to the visual embeddings. The resulting \emph{entropy-gradient} map highlights regions whose visual features most affect the model’s uncertainty, producing a grounding signal that is tightly coupled to the model’s prediction behavior—without auxiliary detectors or attention heuristics.

Crucially, our uncertainty-driven signal supports \emph{multi-region} localization. We convert the entropy-gradient response into multiple spatially coherent regions of interest, enabling explicit aggregation over spatially disjoint evidence. To further recover fine-grained or secondary cues that can be suppressed in a single-pass map by dominant activations, we introduce an iterative refinement procedure that reapplies the same entropy-backpropagation mechanism on selected subregions. We regulate refinement with a spatial-entropy stopping criterion, terminating when spatial concentration no longer improves; empirically, probability-based confidence measures are inconsistent for this control, while spatial entropy provides a stable convergence signal across backbones and tasks.

We evaluate our approach on multiple benchmarks~\cite{mathew2021infographicvqa,wu2023vguidedvisualsearch,mathew2021docvqadatasetvqadocument,xai2024realworldqa,singh2019towards,Li-hallucination-2023,8953451} using strong pretrained backbones~\cite{liu2024improvedbaselinesvisualinstruction, liu2024llavanext, wang2025internvl3_5,Qwen2.5-VL }. Our method consistently improves over these baselines and outperforms the strongest training-free competitor, with gains that are particularly pronounced on detail-critical and multi-evidence queries. Qualitative results further show that our refinement yields more precise and interpretable evidence localization.

Our contributions are summarized as follows:
\begin{itemize}
    \item We introduce a training-free, model-intrinsic grounding method for pretrained VLMs that treats inference as test-time evidence acquisition, using the model’s own uncertainty to guide where to look.
    
    \item We develop a multi-region selection and ranking pipeline that extracts multiple coherent ROIs from entropy-gradient maps, enabling retrieval and aggregation of spatially disjoint evidence required by documents and compositional queries.
    
    \item We introduce an iterative refinement procedure that repeatedly re-grounds and re-crops selected regions, together with a spatial-entropy stopping criterion that provides early stopping.
    
    \item We provide extensive quantitative and qualitative evaluations across multiple benchmarks and demonstrate consistent improvements over baselines.
\end{itemize}

%% file: sec/2_RelatedWorks.tex
\section{Related Work}

\paragraph{\textbf{Visual Grounding in MLLMs.}}
Visual grounding aims to localize image regions relevant to a textual query. Classical approaches rely on supervised detectors, \textit{e.g.,} YOLO~\cite{redmon2016lookonceunifiedrealtime}, or open-vocabulary detectors that condition directly on text, \textit{e.g.,} Grounding DINO~\cite{liu2024groundingdinomarryingdino}. Only limited work studies \emph{training-free} grounding directly from pretrained MLLMs. ViCrop~\cite{zhang2025mllmsknowlooktrainingfree} extracts localization cues from attention-derived signals across forward passes, while other analyses probe attention heads and use heuristic criteria, \textit{e.g.,} spatial entropy, to select informative heads~\cite{kang2025largevisionlanguagemodelneeds}. Several methods instead introduce extra components to convert internal signals into masks, such as F-LMM~\cite{11091875} and LISA~\cite{Lai_2024_CVPR}. More generally, gradient-based attribution methods, \textit{e.g.,} Grad-CAM~\cite{selvaraju2017grad}, Integrated Gradients~\cite{pmlr-v70-sundararajan17a}, indicate that prediction gradients encode spatial saliency, but they are primarily designed for post-hoc interpretability. In contrast, we derive grounding from gradients of an entropy-based objective over the language model’s output distribution, yielding a training-free, model-intrinsic signal that enables principled \emph{multi-region} localization without architectural modification.\vspace{-5pt}

\paragraph{\textbf{Resolution Constraints and Region-Level Reasoning.}}
Reasoning over fine-grained visual evidence remains challenging for MLLMs due to limited spatial granularity and token budgets. Many systems encode the image with fixed-resolution pipelines, e.g., LLaVA-1.5~\cite{liu2024improvedbaselinesvisualinstruction}, while others employ multi-crop or tiling strategies to increase visual coverage, e.g., LLaVA-1.6~\cite{liu2024llavanext}, InternVL~\cite{wang2025internvl3_5}, Molmo~\cite{molmo2024}. Qwen-VL processes images at native resolution without predefined cropping~\cite{Qwen-VL}, and LLaVA-OneVision~\cite{an2025llavaonevision15fullyopenframework} explores native-resolution encoding via RiCe-ViT~\cite{xie2025region,lillava}. Beyond static resolution strategies, several works introduce explicit region-level reasoning. TEVA~\cite{Jiang_2025_ICCV} predicts regions of interest using auxiliary detectors and additional training stages to guide patch selection and visual encoding. SEAL~\cite{wu2023vguidedvisualsearch} performs LLM-guided visual search with dedicated control components and maintains a visual working memory.  We also equip region-level reasoning to existing MLLMs by providing additional image crops discovered through our proposed training-free entropy-based visual grounding method.

%% file: sec/3_Methods.tex
\section{Method}

\subsection{Problem Formulation and Overview}

We consider a pretrained Multimodal Large Language Model (MLLM) that answers a query given an image--prompt pair $(I, P)$. Our goal is to improve visual understanding via \emph{training-free, model-intrinsic} grounding: for each $(I, P)$, we seek to localize the spatial regions in $I$ that constitute the most decision-relevant evidence for answering the query.

\begin{figure}[t]
    \centering
    \includegraphics[width=1.0\linewidth]{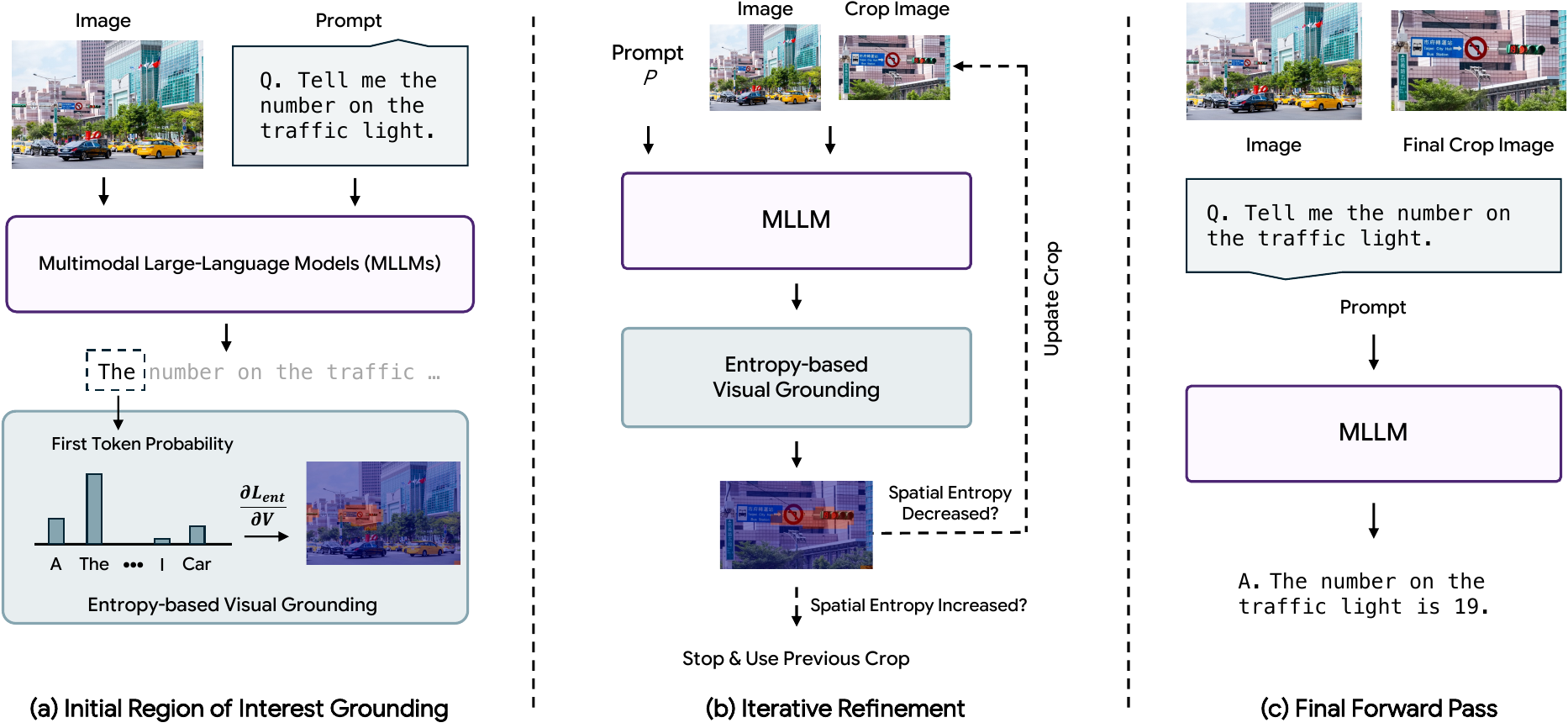}
    \caption{\textbf{Overview of proposed method.} (a) Given an image and prompt, we obtain an initial region-of-interest by backpropagating the entropy of the next-token distribution to visual embeddings, producing an entropy-gradient relevance map. 
(b) We iteratively re-ground and re-crop the most informative regions, stopping when the spatial entropy criterion indicates further refinement no longer improves spatial concentration. 
(c) The final crop(s) are used for a forward pass to produce the answer.}\vspace{-10pt}
    \label{fig:overview}
\end{figure}

In the following, we first describe our entropy-based visual grounding method, which backpropagates the entropy of the next-token distribution to obtain an entropy-gradient map (Sec.~\ref{sec:entropy_grounding}). Since a single saliency map often collapses to a dominant region and fails to capture spatially disjoint evidence, we then introduce a multi-region selection and ranking procedure that extracts multiple coherent regions of interest from the gradient map (Sec.~\ref{sec:multi_region}). Finally, to recover fine-grained or secondary cues and to provide a controlled feedback loop, we present an iterative refinement scheme regulated by a spatial-entropy stopping criterion (Sec.~\ref{sec:iterative_refinement}). An overview of our method is illustrated in Fig.~\ref{fig:overview}.

\subsection{Entropy-Based Visual Grounding}
\label{sec:entropy_grounding}

\begin{figure}[t]
    \centering
    \includegraphics[width=\linewidth]{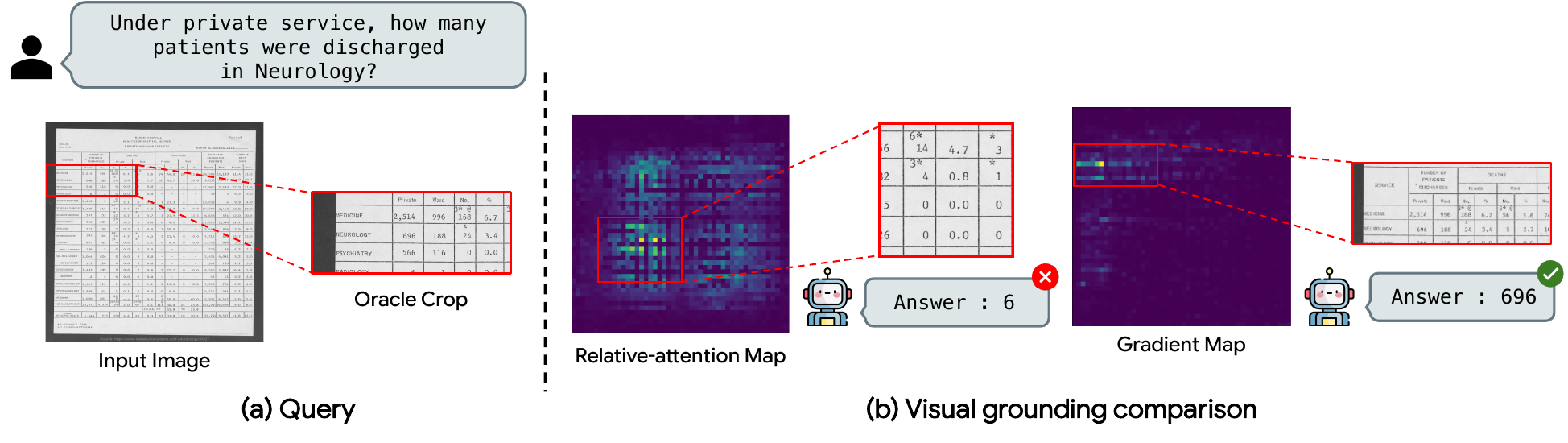}
\caption{\textbf{Relative attention vs. entropy-gradient.} Given an image and a user query as shown in~(a), we compare the attention-based grounding used in ViCrop~\cite{zhang2025mllmsknowlooktrainingfree} with our proposed entropy-gradient grounding in~(b). The relative-attention map produces a diffuse heatmap that highlights an incorrect region, leading to a wrong crop and an erroneous answer (\textcolor{red}{6}). In contrast, the entropy-gradient map concentrates on the query-relevant row of the table, yielding a precise crop that enables the model to correctly answer the question~(\textcolor[HTML]{0B8842}{696}).} \vspace{-10pt}
    \label{fig:attn_grad}
\end{figure}

Our goal is to obtain a {model-intrinsic} grounding signal that reflects {decision-relevant evidence} for a given query. Prior works have explored text-to-image token attention maps to explain which visual regions the model attends to when generating a response~\cite{kaduri2025whats, zhang2025cross,zhang2025mllmsknowlooktrainingfree, kang2025largevisionlanguagemodelneeds,kim2025seg4diff}. However, extracting a clear, interpretable attention map typically requires heuristic selection of specific heads or layers, and often additional post-processing steps~\cite{zhang2025mllmsknowlooktrainingfree,kang2025largevisionlanguagemodelneeds}. Even then, as exemplified in Fig.~\ref{fig:attn_grad}-(b), the resulting maps may be inaccurate, reflect internal token routing rather than the visual evidence that actually drives the model's output, and often the selected attention layers and heads are model-specific, requiring independent analysis for each model. 

\noindent\textbf{Entropy as a grounding objective.}
Rather than aggregating signals spread across many attention operations, we derive grounding directly from the model’s final prediction: its \textit{next-token distribution}. We do so via gradient-based attribution~\cite{selvaraju2017grad}: we define a scalar objective on the output distribution and measure its sensitivity to the visual embeddings. Concretely, we use the Shannon entropy of the next-token distribution and backpropagate it to the visual embeddings, producing an \emph{entropy-gradient} map that is model-intrinsic, query-conditioned, and training-free—requiring no head selection, auxiliary modules, or post-processing heuristics.

Formally, given an image $I$ and prompt $P$, at decoding step $t$ the model produces a next-token distribution $p_t(y)=p(y_t=y\mid I,P,y_{<t})$ over the vocabulary $\mathcal{Y}$. We define its Shannon entropy as
\begin{equation}
H_t(I,P)
\;=\;
-\sum_{y\in\mathcal{Y}} p_t(y)\,\log p_t(y).
\end{equation}

\begin{figure}[t]
\centering
    \includegraphics[width=\linewidth, height=0.55\textheight, keepaspectratio]{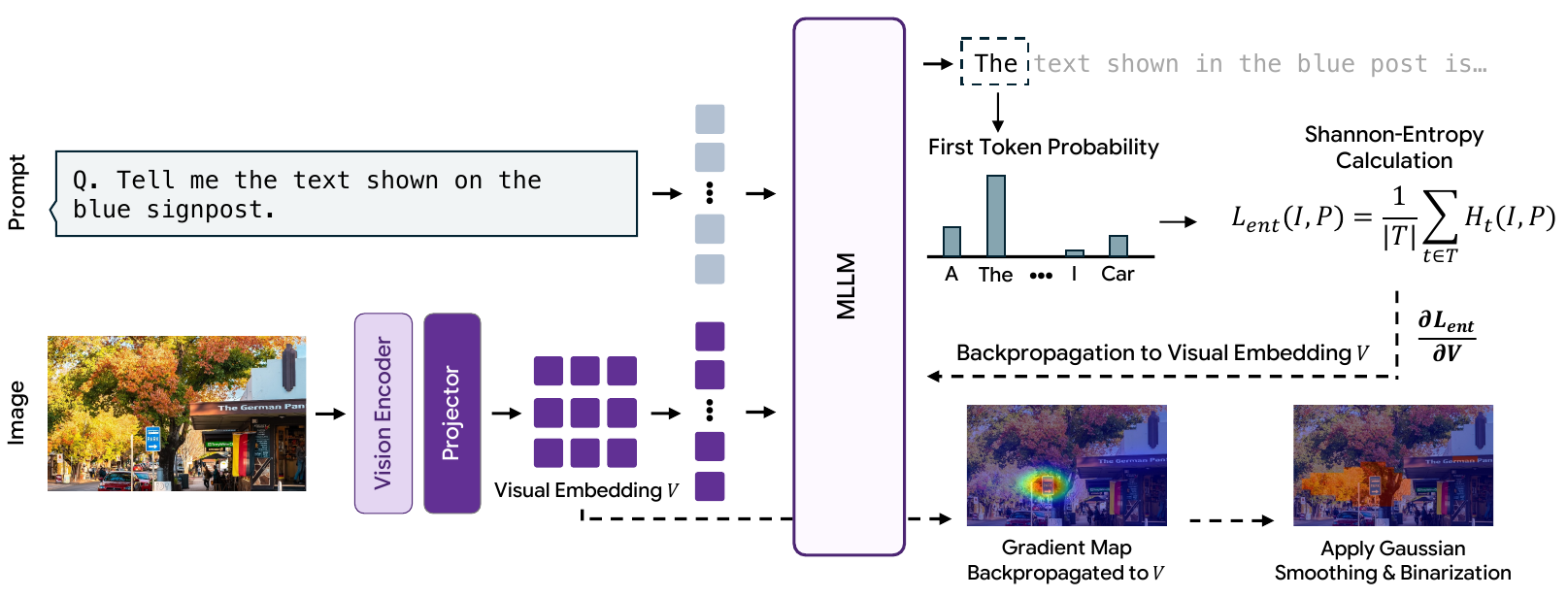}
    \caption{\textbf{Visual grounding via entropy-gradient map.} Given an image and a text prompt, the image is encoded by a vision encoder and projected into visual embeddings V, which are fed alongside the prompt tokens into the MLLM. We compute the Shannon entropy from the next-token probability of the first token. This entropy objective is then backpropagated to the visual embeddings $V$. The resulting gradient map highlights image regions whose visual features most influence the model's predictive uncertainty. A Gaussian smoothing step followed by adaptive binarization converts the continuous map into a clean spatial mask, from which coherent regions of interest are extracted for downstream cropping and answer generation.}\vspace{-10pt}
    \label{fig:grad}
\end{figure}

We use the entropy at a chosen decoding step $t$ as our grounding objective, with $t{=}1$ corresponding to the first decoding step, 
\begin{equation}
\mathcal{L}_{\mathrm{ent}}(I,P)
\;=\;
H_t(I,P),
\end{equation}
and backpropagate it to the projected visual embeddings $\mathbf{V} = \{ \mathbf{v}_1, \dots, \mathbf{v}_N \}$, i.e., the image tokens after the vision-language projector, where $N$ denotes the number of image tokens. We now obtain the entropy-gradient signal as
\begin{equation}
\mathbf{G}
\;=\;
\frac{\partial \mathcal{L}_{\mathrm{ent}}}{\partial \mathbf{V}}.
\end{equation}
Intuitively, $\mathbf{G}$ measures how perturbations to each visual token would change the model’s uncertainty. Since each token $\mathbf{v}_i$ corresponds to a spatial patch, we convert token-wise gradients into scalar saliency scores via the $\ell_2$ norm,
\begin{equation}
s_i \;=\; \left\lVert \mathbf{G}_i \right\rVert_2,
\end{equation}
yielding a score map $\mathbf{S}=\{s_1,\dots,s_N\}$ that can be reshaped into an image-aligned heatmap. The resulting entropy-gradient map highlights regions whose visual evidence most affects the model’s uncertainty about what to generate, serving as the primitive for multi-region extraction and the controlled refinement loop in the following sections.

\subsection{Multi-Region Selection and Ranking from Gradient Maps}
\label{sec:multi_region}

The entropy-gradient score map $\mathbf{S}$ from Sec.~\ref{sec:entropy_grounding} is defined on the visual-token grid and can contain small spurious peaks due to tokenization artifacts and backpropagation noise. Similar to~\cite{kang2025largevisionlanguagemodelneeds}, to obtain stable and spatially coherent regions of interest, we first smooth $\mathbf{S}$ with a Gaussian filter, producing a denoised map $\tilde{\mathbf{S}}$.

We then convert $\tilde{\mathbf{S}}$ into a binary support mask using an adaptive, data-driven threshold. Specifically, we sort the values of $\tilde{\mathbf{S}}$ and apply the elbow  method~\cite{satopaa2011finding} to select a threshold $\tau$ at the point of maximal deviation from the chord connecting the minimum and maximum values. This yields a parameter-free threshold that separates salient responses from background without requiring a manually tuned percentile. With the indicator function $\mathbb{I}[\cdot]$, the binary mask is computed as follows:
\begin{equation}
\mathbf{M}_i \;=\; \mathbb{I}\!\left[\tilde{s}_i \ge \tau\right].
\end{equation}

\begin{figure}[t]
\centering
    \includegraphics[width=\linewidth]{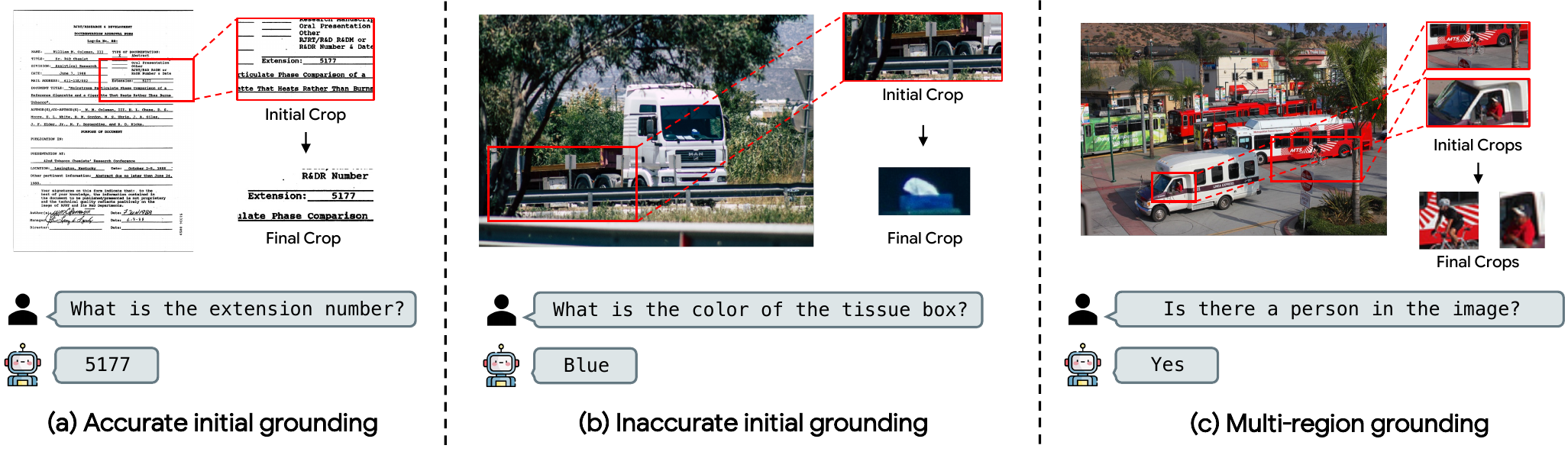}
  \caption{\textbf{Qualitative examples of iterative and multi-region grounding.}
(a) When the initial entropy-gradient crop already covers the evidence (e.g., a small document field), iterative refinement further zooms in to obtain a precise final crop and the correct answer. (b) Even when the initial crop is inaccurate, repeated re-grounding and refinement can recover and converge to the correct evidence region. (c) For queries requiring spatially disjoint cues, our method selects multiple regions of interest and refines them, enabling accurate prediction from multi-region evidence.}\vspace{-5pt}
    \label{fig:comparison}
\end{figure}

The resulting mask typically contains multiple disconnected components, each corresponding to a candidate region of interest. We extract connected components $\{C_j\}$ on the token grid and assign each component an importance score by accumulating the \emph{original} saliency within the region:
\begin{equation}
w_j \;=\; \sum_{i \in C_j} s_i.
\end{equation}
We rank components by $w_j$ and retain the top-$K$ regions. Each selected component is mapped back to the image as a tight bounding box, yielding a compact set of high-evidence, spatially coherent ROIs. These regions are then concatenated with the global view and used as input for final answer generation. The overall pipeline of acquiring binarized gradient maps is illustrated in Fig.~\ref{fig:grad}.

\subsection{Iterative Refinement}
\label{sec:iterative_refinement}

While entropy-based grounding can identify query-relevant regions, a single grounding pass is often insufficient in practice, as exemplified in Fig.~\ref{fig:comparison}. To address these issues, we introduce an iterative refinement procedure that repeatedly re-applies entropy-based grounding and region extraction on the current set of views. We initialize the view set with the original image, which is kept as a global context view, and the top-$K$ regions from Sec.~\ref{sec:multi_region}, ordered by their region scores. At each iteration, we run entropy-based grounding on {each} view, obtain a saliency map and a binarized mask, compute a tight bounding box around the activated support, and crop the corresponding patch. The cropped patches form the view set for the next iteration. This yields a simple test-time~\cite{hong2021deep,hong2024unifying2,hong2022neural} feedback loop: grounding proposes where to look next, and re-invoking the model on these views enables either (i) deeper zoom-in for more decisive evidence or (ii) discovery of alternative regions that were overlooked in earlier iterations.

A key challenge is deciding when further refinement stops helping. We regulate the loop using \emph{spatial entropy}~\cite{kang2025largevisionlanguagemodelneeds}. Given a mask $\mathbf{M}$ with connected components $\{C_i\}_{i=1}^{N}$, we define
\begin{equation}
H(\mathbf{M}) = - \sum_{i=1}^{N} P(C_i)\log P(C_i),
\end{equation}
where
\begin{equation}
P(C_i) = \frac{|C_i|}{\sum_{j=1}^{N} |C_j|},
\end{equation}
and $|C_i|$ denotes the number of active locations in component $C_i$. Spatial entropy measures dispersion: lower values indicate concentrated activation, while higher values indicate diffuse responses. In practice, we track the spatial entropy associated with the most important view. We continue refinement while this entropy decreases, and stop when it increases, indicating that further cropping no longer improves concentration and may start discarding useful context.

%% file: sec/4_Experiments.tex
\section{Experiments}

\subsection{Experimental Settings}

\noindent\textbf{Baselines.} We evaluate our approach on four widely used VLM architectures: LLaVA-1.5 7B~\cite{liu2024improvedbaselinesvisualinstruction}, LLaVA-1.6 Mistral 7B~\cite{liu2024llavanext}, InternVL-3.5 8B~\cite{wang2025internvl3_5}, and Qwen2.5-VL 7B~\cite{Qwen2.5-VL}. 
For comparison, we include TEVA~\cite{Jiang_2025_ICCV}, a training-based grounding model, as well as SEAL~\cite{wu2023vguidedvisualsearch}. We further evaluate the vanilla VLM backbones without our approach and compare against ViCrop~\cite{zhang2025mllmsknowlooktrainingfree} applied to LLaVA-1.5 and LLaVA-1.6. Additional evaluation details for these baselines are provided in the supplementary material.

\noindent\textbf{Datasets.} We evaluate our method on seven VQA benchmarks that stress \emph{fine-grained and multi-region understanding}: TextVQA~\cite{singh2019towards}, V*~\cite{wu2023vguidedvisualsearch}, DocVQA~\cite{mathew2021docvqadatasetvqadocument}, POPE~\cite{Li-hallucination-2023}, InfoQA~\cite{mathew2021infographicvqa}, and benchmarks that evaluate \emph{general and real-world understanding}: GQA~\cite{8953451} and RWQA~\cite{xai2024realworldqa} to verify that our method does not harm general understanding capabilities. Together, they cover scene-text VQA, high-resolution visual search, document and infographic reasoning, compositional visual reasoning, hallucination probing, and real-world spatial understanding. 

\newcommand{\rot}[1]{\rotatebox[origin=lb]{90}{\scriptsize{#1}}}

\begin{table*}[t]
\centering
\caption{\textbf{Quantitative results on standard reasoning benchmarks.} 
Our training-free method improves fine-grained image understanding tasks across four VLM architectures. 
We include TEVA~\cite{Jiang_2025_ICCV} and SEAL~\cite{wu2023vguidedvisualsearch} as training-based references. Results for SEAL are taken from~\cite{zhang2025mllmsknowlooktrainingfree}.}
\label{tab:main_results}
\resizebox{\textwidth}{!}{%
\begin{tabular}{@{}ll|ccccc|cc@{}}
&
  & \rot{TextVQA~\cite{singh2019towards}} & \rot{V*~\cite{wu2023vguidedvisualsearch}} & \rot{DocVQA~\cite{mathew2021docvqadatasetvqadocument}} & \rot{POPE~\cite{Li-hallucination-2023}} & \rot{InfoQA~\cite{mathew2021infographicvqa}}
  & \rot{GQA~\cite{8953451}} & \rot{RWQA~\cite{xai2024realworldqa}} \\ 
  
\cmidrule(lr){1-2}\cmidrule(lr){3-7}\cmidrule(lr){8-9}
\textit{\small VLM} & \textit{\small Method} & \multicolumn{5}{c|}{\textit{\small Fine-grained image understanding}} & \multicolumn{2}{c}{\textit{\small General QA}} \\
\midrule
\multicolumn{9}{l}{\cellcolor[HTML]{FFF3CD}\textit{\textbf{\small Training-based}}} \\[1pt]
TEVA & TEVA 3B\cite{Jiang_2025_ICCV} & 66.80 & 61.60 & - & 87.00 & - & - & - \\
     & TEVA 7B\cite{Jiang_2025_ICCV} & 72.50 & 77.10 & 51.3 & 87.90 & 31.90 & - & - \\

Llava 1.5 & SEAL \cite{wu2023vguidedvisualsearch} & 36.30 & 75.30 & 5.31 & 82.40 & - & 50.18 & - \\

\midrule\midrule
\multicolumn{9}{l}{\cellcolor[HTML]{D9EAD3}\textit{\textbf{\small Training-free}}} \\[1pt]
LLaVA 1.5~\cite{liu2024improvedbaselinesvisualinstruction} & Base model & 46.22 & 46.07 & 22.32 & 86.55 & 22.24 & \textbf{61.98} & \textbf{48.76} \\
          & + ViCrop~\cite{zhang2025mllmsknowlooktrainingfree} & \textbf{55.17} & 47.64 & 19.63 & 87.25 & \textbf{23.26} & 60.97 & 47.97 \\
          & \cellcolor[HTML]{E6E2E2}+ Ours & \cellcolor[HTML]{E6E2E2}52.78 & \cellcolor[HTML]{E6E2E2}\textbf{56.02} & \cellcolor[HTML]{E6E2E2}\textbf{33.70} & \cellcolor[HTML]{E6E2E2}\textbf{87.56} & \cellcolor[HTML]{E6E2E2}22.33 & \cellcolor[HTML]{E6E2E2}61.15 & \cellcolor[HTML]{E6E2E2}48.24 \\
          & \cellcolor[HTML]{E6E2E2} & \cellcolor[HTML]{E6E2E2}{\color[HTML]{009901}+6.56} & \cellcolor[HTML]{E6E2E2}{\color[HTML]{009901}+9.95} & \cellcolor[HTML]{E6E2E2}{\color[HTML]{009901}+11.38} & \cellcolor[HTML]{E6E2E2}{\color[HTML]{009901}+1.01} & \cellcolor[HTML]{E6E2E2}{\color[HTML]{009901}+0.09} & \cellcolor[HTML]{E6E2E2}{\color[HTML]{656565}-0.76} & \cellcolor[HTML]{E6E2E2}{\color[HTML]{656565}-0.52} \\ \midrule
          & Base model & 65.8 & 57.59 & 64.94 & 87.8 & 24.66 & 64.14 & 58.30 \\
          & + ViCrop~\cite{zhang2025mllmsknowlooktrainingfree} & \textbf{68.65} & 61.78 & 51.42 & 88.18 & 28.18 & \textbf{64.54} & 56.99 \\
\multirow{-3}{*}{\shortstack{LLaVA 1.6~\cite{liu2024llavanext}}} & \cellcolor[HTML]{E6E2E2}+ Ours & \cellcolor[HTML]{E6E2E2}67.96 & \cellcolor[HTML]{E6E2E2}\textbf{73.3} & \cellcolor[HTML]{E6E2E2}\textbf{65.07} & \cellcolor[HTML]{E6E2E2}\textbf{89.31} & \cellcolor[HTML]{E6E2E2}\textbf{33.93} & \cellcolor[HTML]{E6E2E2}63.97 & \cellcolor[HTML]{E6E2E2}\textbf{60.39} \\
          & \cellcolor[HTML]{E6E2E2} & \cellcolor[HTML]{E6E2E2}{\color[HTML]{009901}+2.16} & \cellcolor[HTML]{E6E2E2}{\color[HTML]{009901}+15.71} & \cellcolor[HTML]{E6E2E2}{\color[HTML]{009901}+0.13} & \cellcolor[HTML]{E6E2E2}{\color[HTML]{009901}+1.51} & \cellcolor[HTML]{E6E2E2}{\color[HTML]{009901}+9.27} & \cellcolor[HTML]{E6E2E2}{\color[HTML]{656565}-0.17} & \cellcolor[HTML]{E6E2E2}{\color[HTML]{009901}+2.09} \\ \midrule
          & Base model & 59.47 & 47.64 & 58.73 & 84.02 & 41.22 & 58.04 & 61.83 \\
          & \cellcolor[HTML]{E6E2E2}+ Ours & \cellcolor[HTML]{E6E2E2}\textbf{74.29} & \cellcolor[HTML]{E6E2E2}\textbf{67.53} & \cellcolor[HTML]{E6E2E2}\textbf{79.54} & \cellcolor[HTML]{E6E2E2}\textbf{86.7} & \cellcolor[HTML]{E6E2E2}\textbf{53.73} & \cellcolor[HTML]{E6E2E2}\textbf{59.01} & \cellcolor[HTML]{E6E2E2}\textbf{64.71} \\
\multirow{-3}{*}{InternVL 3.5~\cite{wang2025internvl3_5}} & \cellcolor[HTML]{E6E2E2} & \cellcolor[HTML]{E6E2E2}{\color[HTML]{009901}+14.82} & \cellcolor[HTML]{E6E2E2}{\color[HTML]{009901}+19.89} & \cellcolor[HTML]{E6E2E2}{\color[HTML]{009901}+20.81} & \cellcolor[HTML]{E6E2E2}{\color[HTML]{009901}+2.69} & \cellcolor[HTML]{E6E2E2}{\color[HTML]{009901}+12.51} & \cellcolor[HTML]{E6E2E2}{\color[HTML]{009901}+0.97} & \cellcolor[HTML]{E6E2E2}{\color[HTML]{009901}+2.88} \\ \midrule
          & Base model & 80.75 & 73.30 & 90.81	 & 87.00 & 69.02 & \textbf{61.01} & \textbf{67.84} \\
          & \cellcolor[HTML]{E6E2E2}+ Ours & \cellcolor[HTML]{E6E2E2}\textbf{81.45} & \cellcolor[HTML]{E6E2E2}\textbf{86.91} & \cellcolor[HTML]{E6E2E2}\textbf{91.16} & \cellcolor[HTML]{E6E2E2}\textbf{88.47} & \cellcolor[HTML]{E6E2E2}\textbf{73.43} & \cellcolor[HTML]{E6E2E2}59.49 & \cellcolor[HTML]{E6E2E2}66.93 \\
\multirow{-3}{*}{Qwen 2.5 VL~\cite{Qwen2.5-VL}} & \cellcolor[HTML]{E6E2E2} & \cellcolor[HTML]{E6E2E2}{\color[HTML]{009901}+0.70} & \cellcolor[HTML]{E6E2E2}{\color[HTML]{009901}+13.61} & \cellcolor[HTML]{E6E2E2}{\color[HTML]{009901}+0.35} & \cellcolor[HTML]{E6E2E2}{\color[HTML]{009901}+1.47} & \cellcolor[HTML]{E6E2E2}{\color[HTML]{009901}+4.41} & \cellcolor[HTML]{E6E2E2}{\color[HTML]{656565}-1.52} & \cellcolor[HTML]{E6E2E2}{\color[HTML]{656565}-0.91} \\ \bottomrule
\end{tabular}%
}
\end{table*}

\noindent\textbf{Evaluation Metrics.}
For TextVQA, we report the standard VQA accuracy metric.\footnote{\url{https://visualqa.org/evaluation.html}}
For DocVQA and InfoQA, we submit predictions to the official evaluation server and report the official ANLS score returned by the test server.
V$^\ast$, GQA, and RWQA are evaluated using their standard top-1 accuracy.
For POPE, we report accuracy averaged across all splits.

\subsection{Experimental Results}
\label{sec:results}
Tab.~\ref{tab:main_results} reports results on fine-grained image understanding~\cite{singh2019towards,wu2023vguidedvisualsearch,mathew2021docvqadatasetvqadocument,mathew2021infographicvqa,Li-hallucination-2023} and general QA~\cite{xai2024realworldqa,8953451}. We also show qualitative results in Fig.~\ref{fig:comparison}. \vspace{5pt}
In addition to downstream reasoning performance, we report quantitative localization metrics in the appendix.

\noindent\textbf{Fine-grained understanding with consistent gains.}
Overall, our training-free grounding yields the most consistent gains on evidence-critical benchmarks where answers depend on small or spatially dispersed cues. We observe the largest gains on V$^\ast$ across all backbones, consistent with V$^\ast$ emphasizing visually crowded and high-resolution cases that benefit from targeted evidence acquisition. We also see considerable improvements in document and infoQA. Meanwhile, performance on general QA benchmarks, which primarily require global-level information, remains comparable, indicating that our method does not sacrifice broad scene understanding.

\noindent\textbf{Comparison with baselines.}
Compared to ViCrop~\cite{zhang2025mllmsknowlooktrainingfree}, which selects a single crop using heuristic scoring and a fixed crop ratio, our multi-region selection is considerably more robust on tasks requiring evidence aggregation from multiple locations. For example, ViCrop results in performance degradation on DocVQA, whereas applying our method brings substantial improvements. Furthermore, while TEVA~\cite{Jiang_2025_ICCV} is a \textit{\textbf{training-based}} method where its performance should be viewed as an upper bound, our training-free approach outperforms TEVA on tasks such as POPE and InfoQA, and reaches comparable results on V$^\ast$ owing to the significant improvements our grounding provides. We also observe that SEAL~\cite{wu2023vguidedvisualsearch} is largely tailored to perform well on the V$^\ast$ totally failing in other datasets such as TextVQA or POPE.

\begin{figure}[t]
    \centering
    \includegraphics[width=\linewidth, height=0.25\textheight, keepaspectratio]{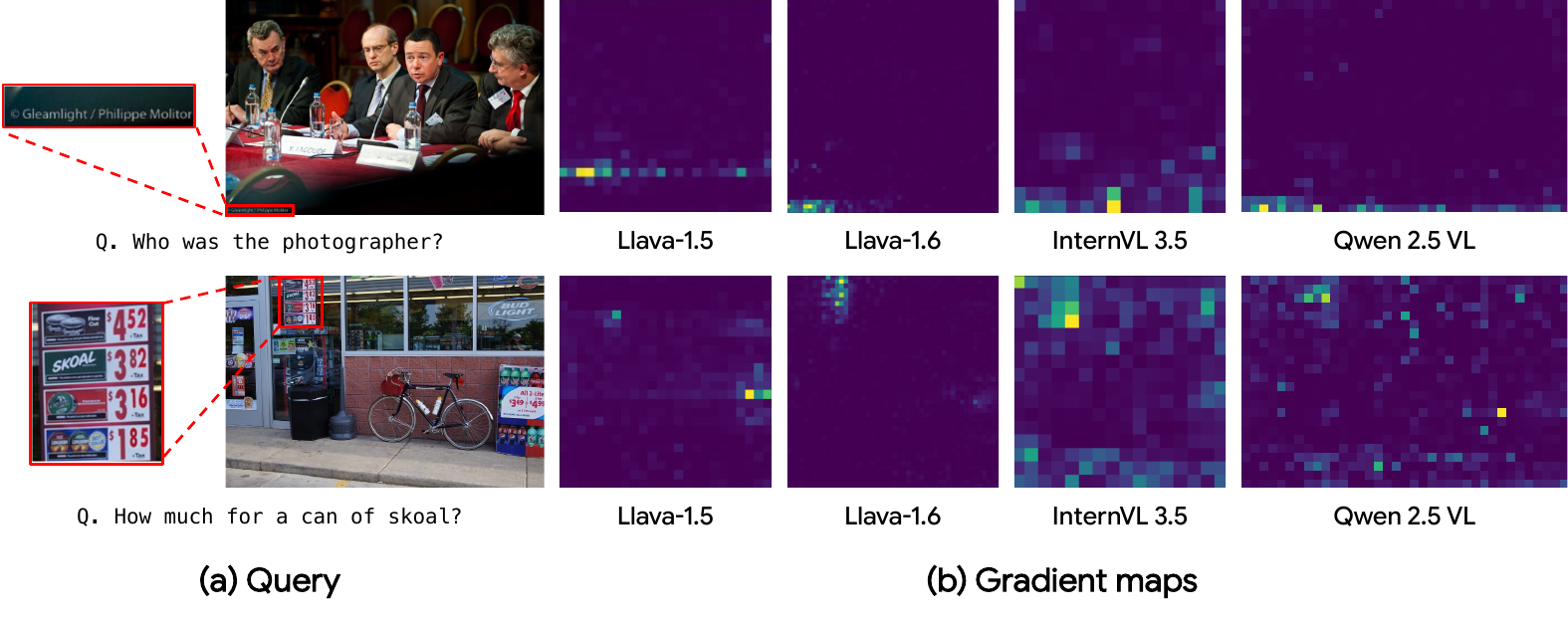}
    \caption{Qualitative comparison of entropy-based gradient maps across different VLMs. We observe that weaker models (e.g., LLaVA 1.5) more frequently allocate gradients to spatially misaligned regions, whereas stronger models produce better-localized attributions. Among the compared models, LLaVA 1.6 generally yields the cleanest and most coherent gradient maps.}\vspace{-10pt}
    \label{fig:entropy_compare}
\end{figure}

\noindent\textbf{Robustness across VLM architectures.}
Our method consistently improves performance across four VLM architectures, suggesting that entropy-gradient grounding generalizes beyond a specific vision encoder or language model. Fig.~\ref{fig:entropy_compare} further shows a qualitative trend: newer/stronger backbones tend to produce more accurately localized entropy-gradient maps, indicating that the quality of our model-intrinsic grounding signal scales with the underlying model’s representations and uncertainty estimates.

\subsection{Ablation Study}
In this section, we present ablation studies and analyses of key design choices. Unless otherwise noted, all experiments are conducted with LLaVA-1.6~\cite{liu2024llavanext}.\vspace{5pt}

\begin{table}[t]
    \centering
    \caption{\textbf{Ablation of different stopping criteria for iterative evidence retrieval.}}\vspace{-5pt}
    \begin{tabular}{l|ccccc}
    \toprule
       Method  & POPE & V* & DocVQA & RWQA & Inference Time \\ \midrule
LLaVA 1.6 Base & 87.8  & 57.59 & 64.94 & 58.30 & \textbf{0.48 s} \\ \midrule\midrule
\multicolumn{6}{c}{\textit{\textbf{\small Threshold-based}}} \\[1pt]
Spatial Entropy &  \textbf{89.31} & \textbf{73.3} & 65.07 & \textbf{60.39} & 4.80 s \\
Confidence     &  89.07 & 70.16 & 64.26  & 58.69 &  3.84 s \\ \midrule\midrule
\multicolumn{6}{c}{\textit{\textbf{\small Iteration-based}}} \\[1pt]
1 Iteration  &  89.30 & 60.73 & \textbf{65.42} & 59.08 &   2.22 s\\
2 Iterations &  89.21 & 65.96 & 63.34  & \textbf{60.39} &  2.70 s \\
3 Iterations &  89.18 & 65.96 & 61.80 & 59.22 & 3.84 s\\ \bottomrule
    \end{tabular}
    \label{tab:merged_ablation}
\end{table}

\noindent\textbf{Stopping criterion and inference cost.} We evaluate two families of stopping criteria for iterative evidence retrieval: \emph{threshold-based} methods, which halt retrieval once a measured signal indicates sufficient evidence, and \emph{iteration-based} methods, which run a fixed number of retrieval steps. For the threshold-based category, we compare our spatial-entropy criterion against a confidence-based alternative that stops when the maximum probability of the first generated token decreases, treating generation confidence as a proxy for evidence sufficiency. Tab.~\ref{tab:merged_ablation} summarizes the results. Among threshold-based methods, spatial entropy provides the best overall accuracy, achieving the highest scores on POPE and V$^\ast$ and matching the best RWQA result, while remaining competitive on DocVQA. The confidence-based heuristic is consistently weaker, suggesting that max-probability is susceptible to decoding instability and does not reliably indicate whether sufficient evidence has been gathered. Fixed iteration budgets reveal that no single iteration count is universally optimal: one iteration slightly improves DocVQA but underperforms on V$^\ast$, while additional iterations can even degrade DocVQA. Although fixed budgets are faster than threshold-based stopping, they require manual tuning per dataset and still fail to match the best accuracy across benchmarks. Overall, spatial entropy offers a robust, adaptive stopping signal with a favorable accuracy--cost trade-off.

\noindent\textbf{Effect of loss function for backpropagation.}
We ablate the objective used to generate gradient-based relevance maps in Tab.\ref{tab:stop_ablation}. In addition to our default entropy objective, we consider two alternatives: (i) a top-$P$ entropy variant that computes entropy over the minimal token set whose cumulative probability mass reaches 90\%, and (ii) a maximum-probability objective that backpropagates the log of the top-1 probability. Overall, performance is stable across objectives, indicating that our localization signal is not overly sensitive to the exact choice of loss. Entropy achieves the best scores on POPE  and V$^\ast$  and ties the best RWQA accuracy, while the top-$p$ entropy slightly improves DocVQA with negligible changes elsewhere. In contrast, the maximum-probability objective is marginally worse on DocVQA  and RWQA, suggesting that relying only on the top-1 token can discard useful distributional information when constructing relevance maps.

\begin{table}[t]
\centering
\caption{\textbf{Ablation of the loss function used for gradient backpropagation.}}\vspace{-5pt}
\label{tab:stop_ablation}
\begin{tabular}{@{}l|cccc@{}}
\toprule
\textbf{Method} & \textbf{POPE} & \textbf{V*} & \textbf{DocVQA} & \textbf{RWQA} \\ \midrule
Entropy             & \textbf{89.31} & \textbf{73.30} & 65.07 & \textbf{60.39}\\
Entropy Top-P       & 89.30          & 72.25          & \textbf{65.09}& \textbf{60.39} \\
Maximum Probability & 89.30          & 73.29          & 64.53    &   60.13   \\ \bottomrule
\end{tabular}
\end{table}

\begin{table}[hbt]
\centering
\caption{\textbf{Effects of varying the number of selected salient regions.}}\vspace{-5pt}
\label{tab:multi_cropst}

\begin{tabular}{@{}l|cccc@{}}
\toprule
\textbf{Method} & \textbf{POPE} & \textbf{V*} & \textbf{DocVQA} & \textbf{RWQA} \\ \midrule
LLaVA 1.6 Base & 87.80 & 57.59 & 64.94 & 58.30 \\ \midrule
1 added & 88.79 & 69.63 & 61.25 & 60.39 \\
2 added & \textbf{89.31} & \textbf{73.30} & 65.07 & \textbf{60.39} \\
3 added & 89.06 & 72.25 & 64.87 & 59.87 \\
4 added & 89.19 & 72.25 & \textbf{65.14} & 58.82 \\ \bottomrule
\end{tabular}
\end{table}

\noindent\textbf{Effect of the number of selected regions.} Tab.~\ref{tab:multi_cropst} ablates the number of appended regions $K$, i.e., how many top-ranked connected components are selected from the gradient map and added as additional views. Adding regions substantially improves fine-grained performance over the vanilla baseline, and we observe a clear optimum at $K{=}2$: performance peaks on POPE and V$^\ast$, while also improving DocVQA and RWQA. While using a single region already boosts V$^\ast$ and RWQA, it can miss complementary evidence and even degrade DocVQA, consistent with document queries often requiring aggregation from multiple locations. 
Increasing $K$ beyond two leads to only marginal and inconsistent changes in performance, suggesting diminishing returns as additional regions may introduce redundancy.
Overall, selecting a small number of regions provides a good trade-off between capturing complementary evidence and avoiding redundant visual crops.

\begin{table}[hbt]
\centering
\caption{\textbf{Effects of different layers for gradient computation.}}\vspace{-5pt}
\label{tab:layer_ablation}
    \begin{tabular}{@{}l|cccc@{}}
    \toprule
    \textbf{Method} & \textbf{POPE} & \textbf{V*} & \textbf{DocVQA} & \textbf{RWQA} \\ \midrule
    Layer 15        & 88.32          & 58.64          & 50.99           & 58.17          \\
    Layer 20        & 88.84          & \textbf{78.01}          & 62.68           & \textbf{61.18}          \\
    Layer 26        & 89.07          & 72.25          & 64.75           & 60.78          \\
    Layer 32 (Last) & \textbf{89.31} & 73.30 & \textbf{65.07}  & 60.39 \\ \bottomrule
\end{tabular}
\end{table}

\begin{figure}[hbt]
    \centering
\includegraphics[width=0.75\linewidth]{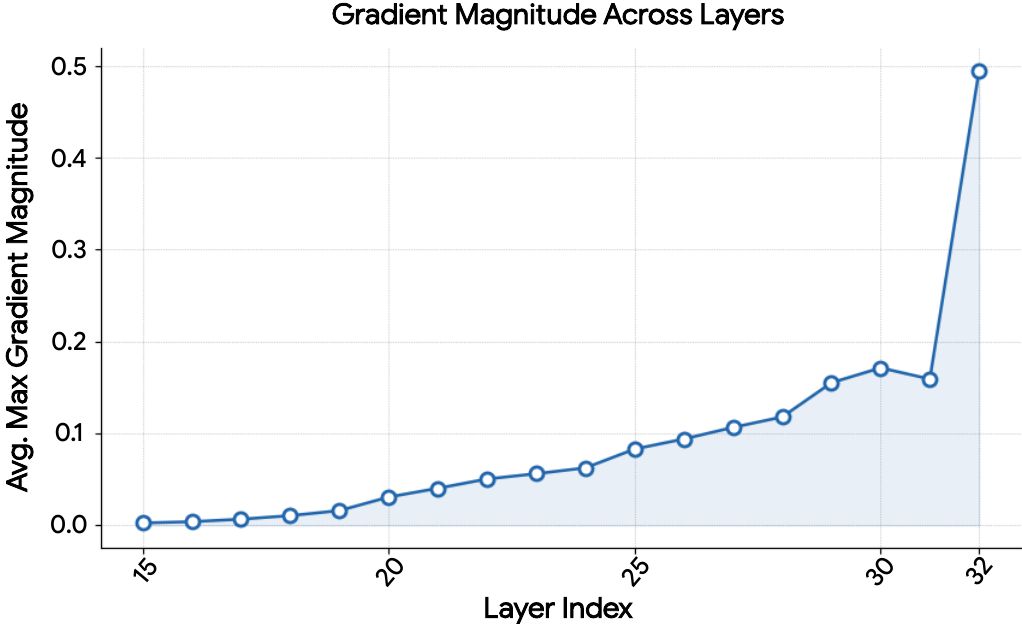}
\caption{Layer-wise maximum average gradient magnitude on TextVQA, illustrating how gradient strength evolves with model depth.}\vspace{-10pt}
    \label{fig:amp2}
\end{figure}

\noindent\textbf{Layer-wise gradient analysis.}
We vary the transformer layer used to compute the entropy-gradient signal and report the performance in Tab.~\ref{tab:layer_ablation}. It shows that shallow layers perform poorly, while deeper layers yield markedly stronger results. The final layer is the most robust overall, consistent with Fig.~\ref{fig:amp2} where gradient magnitudes increase toward deeper layers. We therefore use the last layer in all experiments; earlier layers are omitted due to consistently weak performance.

\begin{table}[t]
\centering
\caption{Performance when backpropagating the loss from different generated tokens. Results are reported on~\cite{mathew2021docvqadatasetvqadocument,singh2019towards}, as these datasets require multi-token generation.}\vspace{-5pt}
\label{tab:second_token}
\begin{tabular}{@{}l|cccc@{}}
\toprule
\textbf{Dataset} & \textbf{First} & \textbf{Second} & \textbf{Third} & \textbf{Fourth} \\ \midrule
TextVQA & \textbf{67.96} & 65.27 & 67.37 & 66.35 \\
DocVQA  & 65.07 & 63.70 & \textbf{65.28} & 64.64 \\
\bottomrule
\end{tabular}
\end{table}

\begin{figure}[hbt]
    \centering
\includegraphics[width=1.0\linewidth]{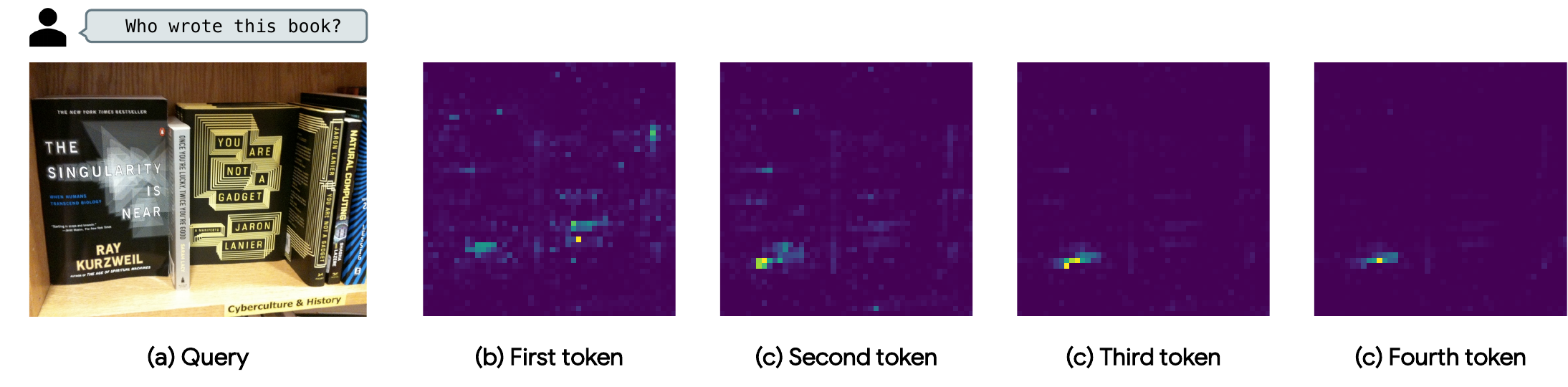}
    \caption{\textbf{Example of gradient computation at different token positions.} Gradients taken at later tokens show stronger conditioning on specific regions, which can lead to better localization but also reduces exploration of other areas.}\vspace{-10pt}
    \label{fig:next_token}
\end{figure}

\noindent\textbf{Subsequent-token loss backpropagation.} We study whether gradients from later decoding steps provide a better grounding signal than those from the first step. Tab.~\ref{tab:second_token} reports results when backpropagating the loss from the $t$-th generated token ($t\in\{1,2,3,4\}$) while keeping all other components fixed. Backpropagating from the first token performs best overall: using later tokens generally degrades TextVQA and does not yield consistent improvements on DocVQA (the third token provides a slight gain, but the trend is not stable across tokens). As visualized in Fig.~\ref{fig:next_token}, gradients from subsequent tokens become increasingly conditioned on previously generated tokens, which is expected under causal self-attention and can bias the relevance map toward an already committed interpretation. In addition, using token $t>1$ requires unrolling multiple decoding steps prior to backpropagation, increasing compute and memory. We therefore use the first token as a simple and efficient choice in all experiments.
\vspace{-10pt}

\noindent\textbf{Computational overhead.}
We report the additional inference time introduced by our method as the average per-sample runtime on RealWorldQA~\cite{xai2024realworldqa} (726–1536 px images), shown in Tab.~\ref{tab:inferenceTime}. We report both a single retrieval pass and the full iterative variant with our stopping criterion. As expected, iterative evidence retrieval increases runtime due to repeated forward/backward passes. Despite this overhead, runtimes remain practical, and the stopping criterion limits unnecessary iterations, yielding a bounded cost while providing the accuracy gains reported in Tab.~\ref{tab:main_results}.

\begin{table}[H]
\centering
\caption{Inference time comparison between a single and iterative runs.}\vspace{-5pt}
\label{tab:inferenceTime}

\begin{tabular}{@{}l|cccc@{}}
\toprule
\textbf{Method} & \textbf{LLaVA-1.5} & \textbf{LLaVA-1.6} & \textbf{InternVL-3.5} & \textbf{QWENVL-2.5} \\ \midrule
Single iteration    & 1.00 s & 2.04 s & 1.18 s & 1.18 s\\
Iterative (w/ stopping) & 3.10 s & 3.37 s & 3.17 s & 2.98 s  \\ \bottomrule
\end{tabular}
\end{table}

%% file: sec/5_Conclusion.tex
\section{Conclusion}
In this work, we introduced a training-free, model-intrinsic visual grounding framework for pretrained VLMs by backpropagating the entropy of the next-token distribution to visual embeddings. Using uncertainty gradients as a decision-relevant signal and converting them into ranked regions of interest, our method retrieves evidence from spatially disjoint cues without auxiliary detectors or heuristic attention processing. To address fixed-resolution limitations, we further propose an iterative refinement loop guided by a spatial-entropy stopping criterion, enabling the model to acquire finer-grained evidence and recover overlooked regions at inference time. Extensive experiments across standard reasoning benchmarks and four VLM architectures show consistent improvements on evidence-critical tasks—particularly in high-resolution and document-centric settings—while producing more focused, query-conditioned localizations.

%% file: sec/supple.tex
\setcounter{page}{1}
\begin{center}
    \Large \bfseries Entropy-Gradient Grounding: Training-Free Evidence Retrieval in Vision-Language Models \\ [5pt]
    \large --Appendix--
\end{center}
\setcounter{tocdepth}{2}
\renewcommand{\contentsname}{\normalsize\textbf{Table of Contents}}

\makeatletter
\renewcommand\tableofcontents{%
  \vspace{10pt}%
  {\noindent\large\textbf{Table of Contents}\par\nobreak\vskip 6pt}%
  \@starttoc{toc}%
}
\makeatother
\providecommand{\authcount}[1]{}

\tableofcontents
\clearpage

\addtocontents{toc}{\protect\setcounter{tocdepth}{2}}
\section{Pytorch-Style Psuedo Code}
\label{appendix:implementation_details}
We provide a Pytorch-Style psuedo code in Alg. \ref{alg:entropy_grounding}.

\begin{algorithm}[hbt]
\caption{Entropy-Driven Multi-Region Localization with Iterative Refinement}
\label{alg:entropy_grounding}
\DontPrintSemicolon
\KwIn{Image $I$, prompt $P$, VLM $\mathcal{M}$, max iterations $T$, top regions $K$}
\KwOut{Refined image set $\mathcal{I}$}

Initialize $\mathcal{I}_0 \leftarrow \{ I \}$\;
Initialize $H_{\text{prev}} \leftarrow +\infty$\;

\For{$t = 0$ \KwTo $T-1$}{
    Initialize candidate component set $\mathcal{C} \leftarrow \emptyset$\;
    
    \ForEach{$I_i \in \mathcal{I}_t$}{
        Compute next-token distribution $p(y \mid I_i, P)$\;
        Compute entropy loss $\mathcal{L}_{\mathrm{ent}}$\;
        Backpropagate $\mathcal{L}_{\mathrm{ent}}$ to visual embeddings\;
        Obtain saliency map $\mathbf{S}_i$\;
        Extract connected components $\{C_{i,j}\}$ from $\mathbf{S}_i$\;
        
        \ForEach{$C_{i,j}$}{
            Compute component score $w_{i,j}$\;
            Add $(I_i, C_{i,j}, w_{i,j})$ to $\mathcal{C}$\;
        }
    }
    
    Select top-$K$ components from $\mathcal{C}$ by score\;
    
    Let $(I^\star, C^\star)$ be the highest-scoring component\;
    Compute spatial entropy $H_t$ of $C^\star$\;
    
    \If{$H_t \geq H_{\text{prev}}$}{
        \textbf{break}\;
    }
    $H_{\text{prev}} \leftarrow H_t$\;
    
    Initialize $\mathcal{I}_{t+1} \leftarrow \{ I \}$ \tcp*{retain global context}
    
    \ForEach{selected $(I_i, C_i)$}{
        Crop bounding box of $C_i$ from $I_i$ to obtain $I_{C_i}$\;
        Add $I_{C_i}$ to $\mathcal{I}_{t+1}$\;
    }
}

\Return{$\mathcal{I}_t$}\;
\end{algorithm}
\clearpage

\section{Evaluation Details}
For each baseline model~\cite{liu2024improvedbaselinesvisualinstruction,liu2024llavanext,Qwen2.5-VL,wang2025internvl3_5}, we retain its default inference configuration and apply our method on top without modifying the baseline itself. For example, LLaVA-1.5 takes a single input image by default, whereas LLaVA-1.6 uses five images. Since our approach naturally accommodates varying numbers of input images, we preserve these settings as is. Similarly, InternVL and Qwen use original-resolution images in their default inference setups, and our method remains fully compatible with this setting, allowing us to integrate it directly without any additional changes.

\section{Dataset Information}
This section briefly summarizes the datasets used in our evaluation.
\begin{itemize}
    \item \textit{TextVQA}~\cite{singh2019towards}: A scene-text VQA dataset where answering questions requires reading and reasoning about text embedded in natural images, making it particularly sensitive to small, detail-critical visual evidence.

    \item \textit{V$^\ast$}~\cite{wu2023vguidedvisualsearch}: A benchmark introduced alongside the V$^\ast$ guided visual search framework to evaluate detail-focused understanding in visually crowded and high-resolution scenes, where relevant cues are often small and easily overlooked.

    \item \textit{DocVQA}~\cite{mathew2021docvqadatasetvqadocument}: A document-image VQA dataset requiring text- and layout-aware reasoning across diverse document types (e.g., forms, receipts, and papers), often involving locating multiple pieces of information within a page.

    \item \textit{InfoQA}~\cite{mathew2021infographicvqa}: A VQA benchmark on infographic images requiring joint reasoning over textual content, layout, graphics, and data visualizations, frequently involving elementary arithmetic and multi-step evidence aggregation.

    \item \textit{GQA}~\cite{8953451}: A large-scale compositional VQA benchmark designed for structured visual reasoning, where questions depend on relationships between multiple objects and attributes.

    \item \textit{POPE}~\cite{Li-hallucination-2023}: A polling-based probing benchmark designed to evaluate \emph{object hallucination} in VLMs using controlled yes/no queries about object presence.

    \item \textit{RWQA}~\cite{xai2024realworldqa}: A real-world reasoning benchmark introduced with Grok-1.5V, consisting of images from everyday and vehicle-captured scenarios with verifiable answers, emphasizing spatial and physical reasoning in real-world environments.
\end{itemize}

Furthermore, TextVQA, InfoQA, and DocVQA provide additional OCR tokens extracted by a third-party model to support inference. In our evaluations, we do not utilize these tokens.

\label{NativeIntern}

\clearpage
\section{Comparison with Additional Methods}

Here, we further compare against ZoomEye~\cite{shen-etal-2025-zoomeye} on \mbox{LLaVA-1.5 7B}, the only backbone shared by both works. Although both approaches are also compatible with Qwen 2.5, ZoomEye is evaluated with the 3B variant, whereas our method uses the 7B variant. In our experiments, running ZoomEye on Qwen 2.5 7B was infeasible due to out-of-memory issues, even on an A100 GPU with 80GB VRAM.

ZoomEye is a training-free, model-agnostic tree-search method that represents an image as a hierarchical tree, where the root denotes the full image and each child node corresponds to a zoomed-in sub-region of its parent. Given a question, it guides the MLLM to traverse this tree by assigning confidence-based priority scores to candidate nodes, effectively mimicking human-like zoom-in behavior to identify task-relevant visual evidence. The search terminates once the model’s answer confidence exceeds a predefined threshold.

\begin{table}[]
\centering
\caption{\textbf{Quantiative comparison to ZoomEye on Llava 1.5.}}
\label{tab:zoomeye_comparison}
\begin{tabular}{@{}l|ccccc|cc@{}}
\toprule
           & TextVQA & V* & DocVQA & POPE & InfoQA & GQA & RWQA  \\ \midrule
Base model & 46.22 & 46.07 & 22.32 & 86.55 & 22.24 & 61.98 & 48.76\\ \midrule
ZoomEye & 46.68 & 72.25    &  24.61     & 86.85   &23.23  &  61.42   & 49.67    \\
Ours       &  52.78        & 56.02   &  33.70      &  87.56     &  22.33       &  61.15   &   48.24  
\end{tabular}
\end{table}

We compare ours with ZoomEye across several benchmarks in Tab.~\ref{tab:zoomeye_comparison}. While ZoomEye achieves strong performance on the V$^*$ benchmark, which consists of high-resolution images with fine-grained visual elements that benefit from its exhaustive tree-based exploration, it does not generalize as effectively to other datasets. In contrast, our method demonstrates more consistent improvements across diverse reasoning tasks, particularly on TextVQA and DocVQA. This pattern is similar to what we observe with SEAL~\cite{wu2023vguidedvisualsearch}: methods that are specifically designed for high-resolution visual search scenarios tend to excel on benchmarks like V$^*$ but may struggle on tasks that require different forms of visual reasoning, such as document understanding or scene-text recognition. Our entropy-gradient grounding, by contrast, provides a more general-purpose grounding signal that adapts to a wider range of query types and visual contexts without relying on task-specific search heuristics.

\clearpage
\section{Quantitative Evaluation of Localization Accuracy}

While the main paper demonstrates qualitative grounding results and quantitative improvements on downstream tasks, it is equally important to directly assess the localization accuracy of our grounding signal. To this end, we measure the Intersection-over-Union (IoU) between the spatial regions identified by our entropy-gradient mask and the ground-truth bounding boxes for TextVQA annotated by ViCrop~\cite{zhang2025mllmsknowlooktrainingfree}. All evaluations are conducted with LLaVA 1.6.

\begin{wraptable}[8]{r}{0.4\linewidth}
    \centering
    \vspace{-30pt}
    \caption{\textbf{Quantitative localization performance on TextVQA using ground-truth bounding boxes provided by ViCrop.}}
    \resizebox{0.6\linewidth}{!}{
    \begin{tabular}{l|cc}
        \toprule
        Metric & Ours & ViCrop \\
        \midrule
        IoU & 0.29 & 0.14 \\
        \bottomrule
        \end{tabular}}
    \label{tab:textvqa_localization}
\end{wraptable}

Tab.~\ref{tab:textvqa_localization} shows that our method achieves substantially higher IoU than ViCrop, indicating more accurate localization of query-relevant visual evidence. This result is consistent with the qualitative comparisons in Fig.~\ref{fig:attn_grad} and Fig.~\ref{fig:comparison}, and provides direct quantitative evidence that backpropagating entropy to visual embeddings yields a grounding signal better aligned with the spatial evidence necessary for answering the question.

\section{Qualitative Comparison with Additional Gradient-based Grounding Methods}

While our method backpropagates the entropy of the next-token distribution to the projected visual embeddings $\mathbf{V}$, alternative choices exist for both the backpropagation target and the objective. For example, one could backpropagate gradients to raw image pixels instead of visual embeddings, or optimize the log-probability of the top-1 token rather than Shannon entropy.

ViCrop~\cite{zhang2025mllmsknowlooktrainingfree} considers such an alternative in its \texttt{pure-grad} ablation. For each image--question pair $(x, q)$, it computes the log of the maximum output probability at the first answer token and visualizes the resulting gradients after taking the $\ell_2$ norm across image channels. Our method differs in two key ways. First, we backpropagate to \emph{visual embeddings} rather than raw pixels, which yields semantically richer gradients that are better aligned with the model’s internal representation space. Second, we optimize the Shannon entropy of the full next-token distribution, which captures the model’s global uncertainty over the vocabulary, rather than the log-probability of a single predicted token, which ignores the rest of the distribution.

\begin{figure}[t]
\centering

\begin{subfigure}{\linewidth}
    \centering
    \includegraphics[width=\linewidth]{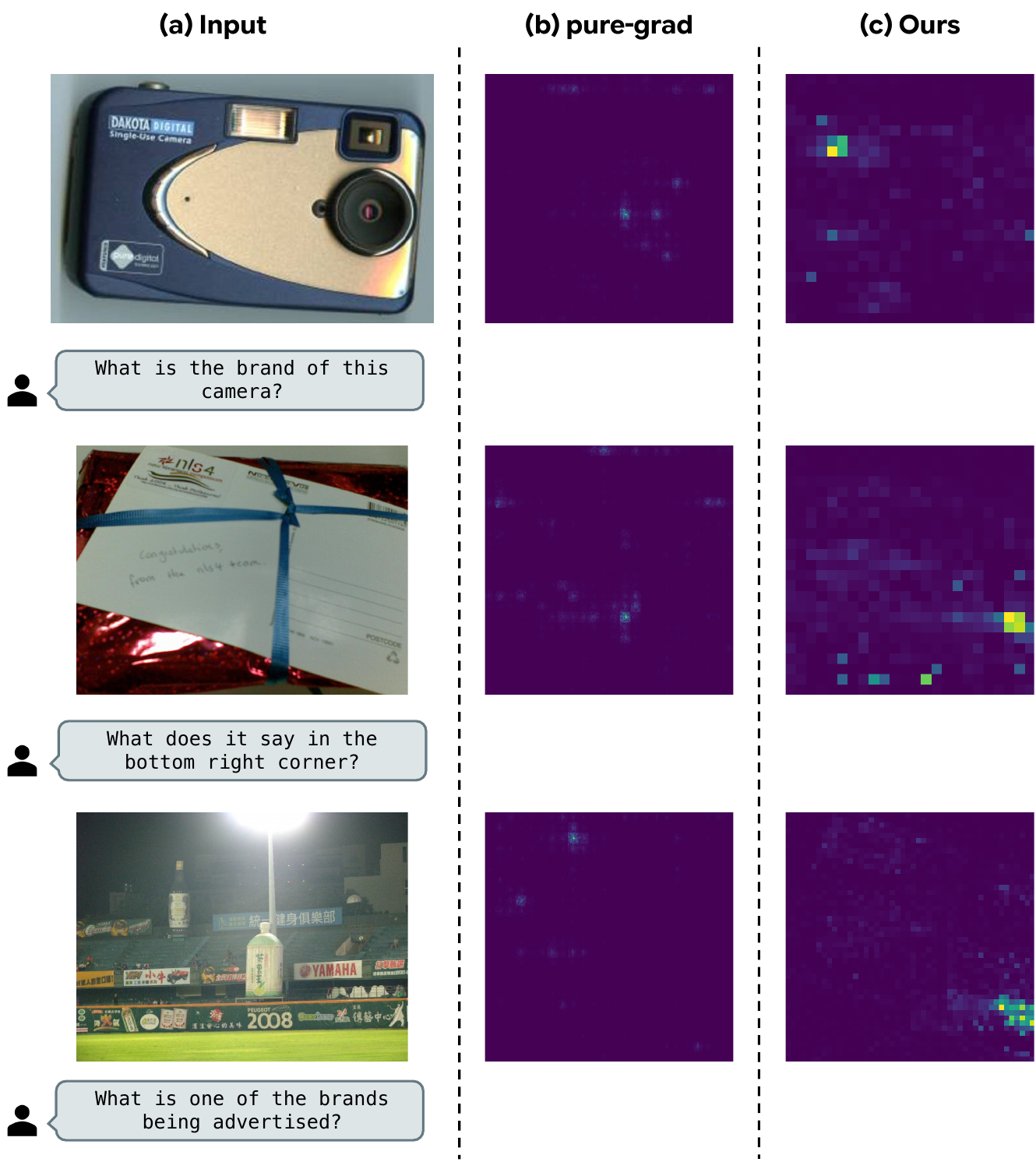}

\end{subfigure}

\caption{\textbf{Qualitative comparison of gradient-based grounding methods on LLaVA~1.5.} For each example, we show (a)~the input image with the corresponding question, (b)~the gradient map produced by the \texttt{pure-grad} method of ViCrop~\cite{zhang2025mllmsknowlooktrainingfree}, which backpropagates the log-probability of the top-1 token to the input pixels, and (c)~the entropy-gradient map produced by our method, which backpropagates the Shannon entropy to the visual embeddings. The \texttt{pure-grad} maps appear diffuse and lack discernible spatial structure, whereas our entropy-gradient maps produce well-localized activations that concentrate on the query-relevant regions, confirming the effectiveness of operating in the embedding space with an entropy-based objective.}
\label{fig:pure_grad_comparison}
\end{figure}

Fig.~\ref{fig:pure_grad_comparison} shows that \texttt{pure-grad} produces diffuse and poorly structured gradient maps, making query-relevant regions difficult to identify. By contrast, our entropy-gradient maps are spatially compact and well localized, clearly highlighting the visual evidence needed to answer the question. These qualitative comparisons support our design choice of backpropagating entropy to visual embeddings as a more effective grounding strategy.

\clearpage
\section{Prompts}

In this section, we list the prompts used during inference for all evaluated models. We report the exact phrasing used for the questions, while the full prompt structure is generated using each model's native prompt template. Among the datasets we use for evaluation, TextVQA, DocVQA, InfoVQA, GQA, and POPE datasets are open-ended questions, where V$^\ast$ and RWQA datasets ask the model to answer from multiple choices. For each category, we use the prompts formatted as follows:
\begin{itemize}
    \item \noindent\textbf{Open-ended questions} 
\end{itemize}
\begin{tcolorbox}[colback=gray!5, colframe=gray!50, fontupper=\ttfamily, title=Prompt Template]
\{Question\}\\
Answer the question using a single word or phrase.
\end{tcolorbox}

\begin{itemize}
    \item \noindent\textbf{Multiple-choice questions} 
\end{itemize}
\begin{tcolorbox}[colback=gray!5, colframe=gray!50, fontupper=\ttfamily, title=Prompt Template]
\{Question\}\\
(A) ... \\
(B) ... \\
(C) ... \\
(D) ... \\
Answer with the option's letter from the given choices directly.
\end{tcolorbox}

\clearpage
\section{Additional Qualitative Examples}

\begin{figure}[hbt]
\centering

\begin{subfigure}{\linewidth}
    \centering
    \includegraphics[width=\linewidth]{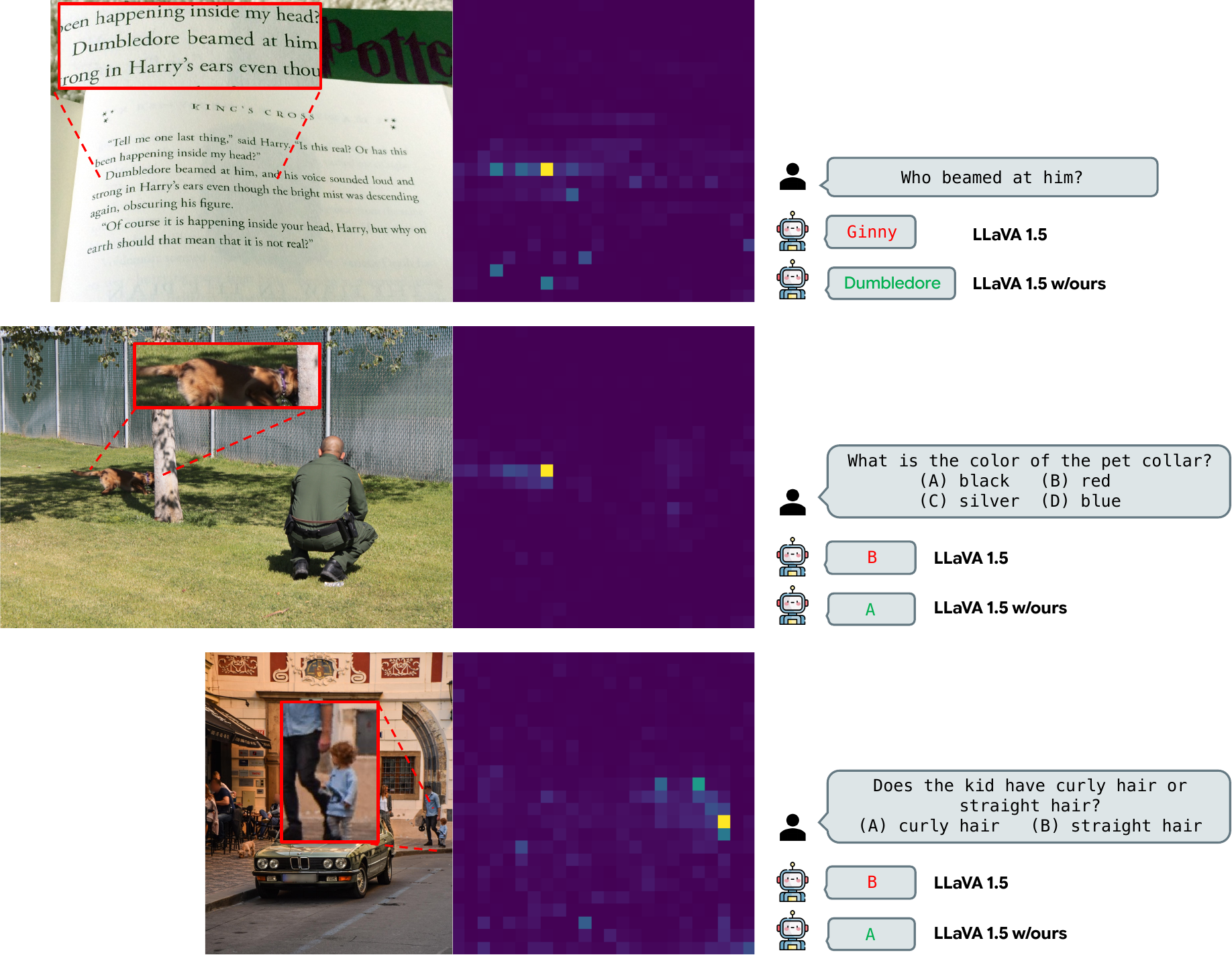}

\end{subfigure}

\caption{\textbf{Additional Examples.} Qualitative examples on Llava 1.5. The most important final crop is highlighted
in red.}
\label{fig:llava15}
\end{figure}

\begin{figure}[t]
\centering

\begin{subfigure}{\linewidth}
    \centering
    \includegraphics[width=\linewidth]{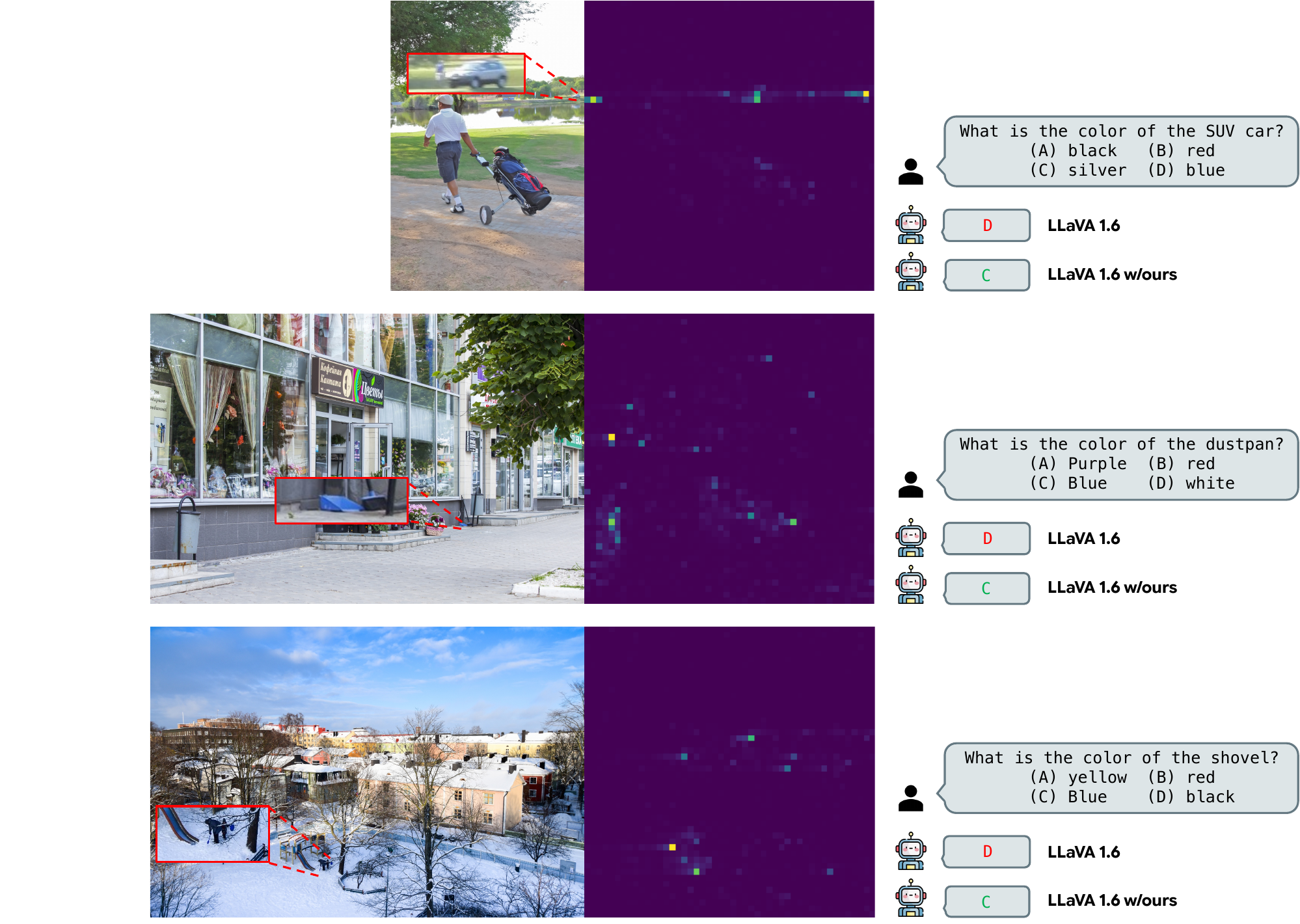}

\end{subfigure}

\caption{\textbf{Additional Examples.} Qualitative examples on Llava 1.6. The most important final crop is highlighted in red.}
\label{fig:llava16}
\end{figure}

\begin{figure}[t]
\centering

\begin{subfigure}{\linewidth}
    \centering
    \includegraphics[width=\linewidth]{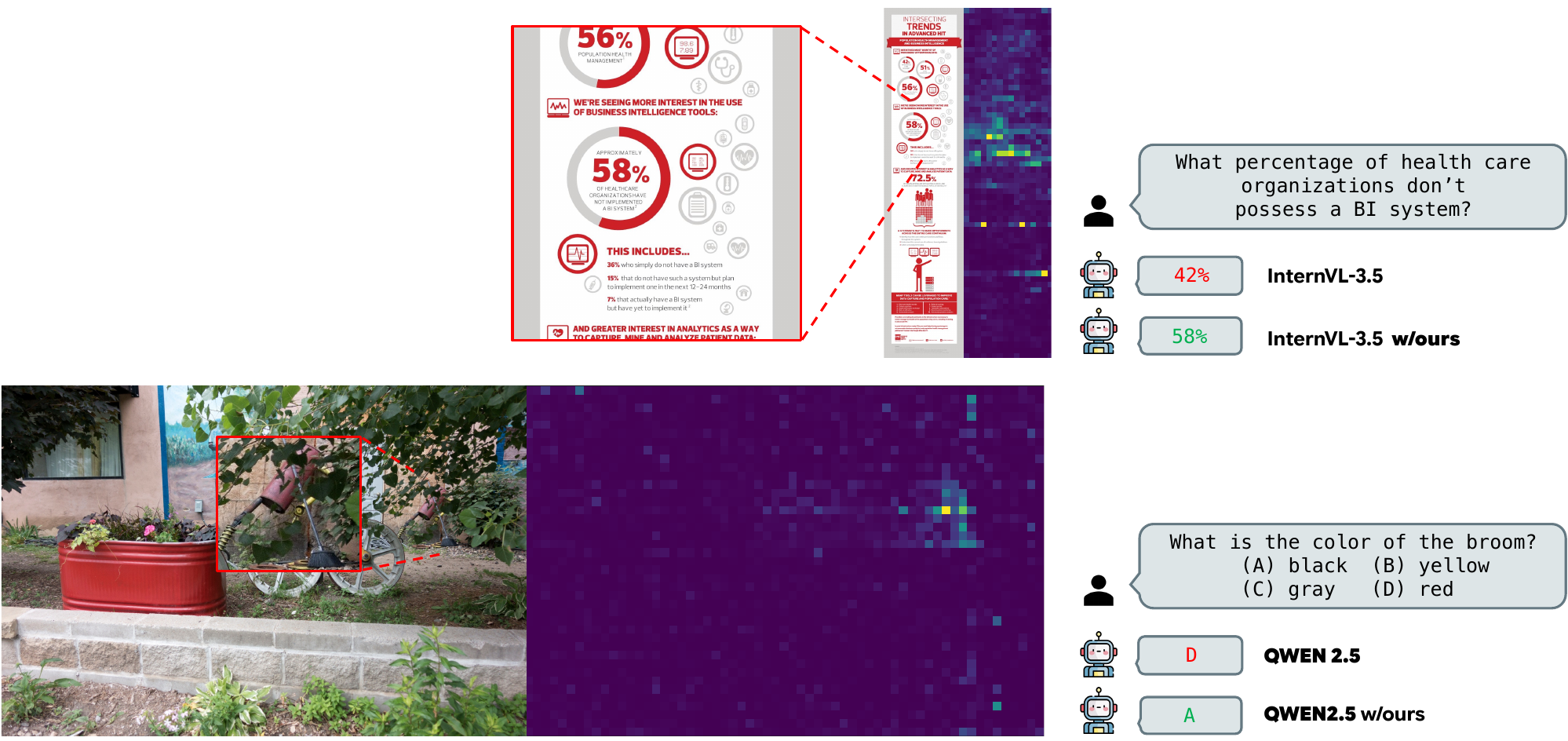}

\end{subfigure}

\caption{\textbf{Additional Examples.} Qualitative examples on native resolution encodings~(QWEN 2.5, InternVL 3.5). The most important final crop is highlighted in red.}
\label{fig:NativeResExample}
\end{figure}

In this section, we provide additional qualitative examples of the entropy-gradient map and the final crop after our interactive refinement applied to various baselines. The examples are shown in Fig.~\ref{fig:llava15}, Fig.~\ref{fig:llava16}, and Fig.~\ref{fig:NativeResExample}.

\clearpage
\section{Limitations}

\begin{figure}[hbt]
\centering

\begin{subfigure}{\linewidth}
    \centering
    \includegraphics[width=\linewidth]{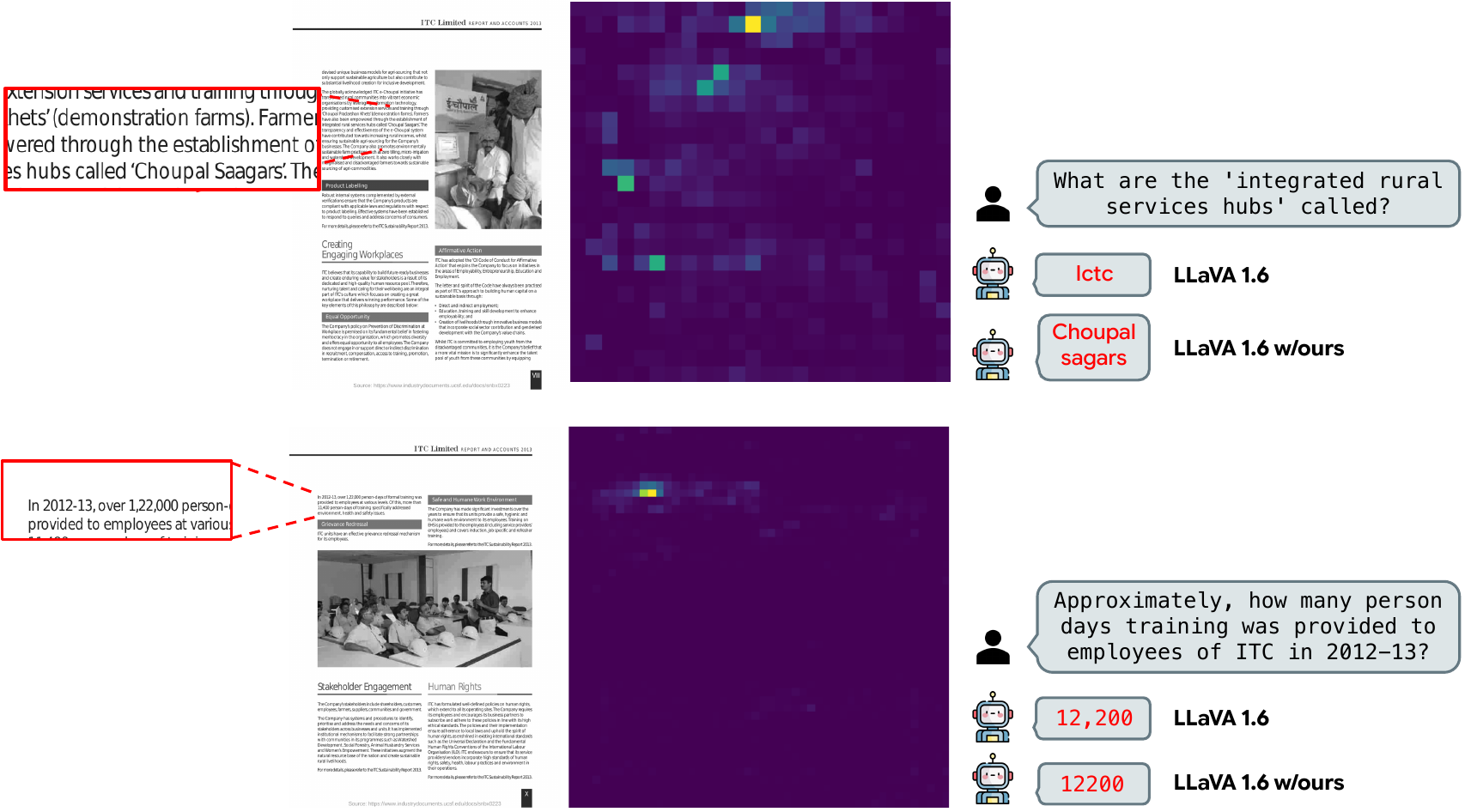}

\end{subfigure}

\caption{\textbf{Failure cases.} Qualitative examples of failing to predict the correct answer illustrating a limitation of our method. Even when confronted with a very detailed crop, the model still fails to answer correctly.}
\label{fig:Limitation1}
\end{figure}

Although our method improves the localization of query-relevant visual evidence, correct localization by itself is not sufficient to ensure a correct final answer. The downstream language model must still accurately interpret the visual content and carry out the required reasoning, and thus errors may persist even when the relevant region is localized correctly. As shown in Fig.~\ref{fig:Limitation1}, the model can still misinterpret a detailed crop that zooms into the correct location to answer the question. Furthermore, our method may fail to recover relevant regions when the backbone model itself does not consider them important, since it ultimately relies on the spatial signals produced by the underlying model (see Fig.~\ref{fig:entropy_compare}).

